# Witscript 3: A Hybrid AI System for Improvising Jokes in a Conversation

**Joe Toplyn**
Twenty Lane Media, LLC
P. O. Box 51
Rye, NY 10580 USA
joetoplyn@twentylanemedia.com

## Abstract

Previous papers presented Witscript and Witscript 2, AI systems for improvising jokes in a conversation. Witscript generates jokes that rely on wordplay, whereas the jokes generated by Witscript 2 rely on common sense. This paper extends that earlier work by presenting Witscript 3, which generates joke candidates using three joke production mechanisms and then selects the best candidate to output. Like Witscript and Witscript 2, Witscript 3 is based on humor algorithms created by an expert comedy writer. Human evaluators judged Witscript 3's responses to input sentences to be jokes 44% of the time. This is evidence that Witscript 3 represents another step toward giving a chatbot a humanlike sense of humor.

## Introduction

Generating all types of humor is often regarded as an AI-complete problem (Winters 2021). But for a conversational agent like a social robot to be truly useful, it must be able to generate contextually integrated jokes about what's happening at the moment (Ritchie 2005). What's more, to be high-quality, a conversational agent must generate a variety of jokes using a variety of joke production mechanisms (Amin and Burghardt 2020). Witscript 3 is a novel system for automatically generating a diverse range of contextually integrated jokes using three different joke production mechanisms.

## Related Work

A few systems for the computational generation of verbally expressed humor can generate contextually integrated jokes, such as jokes improvised in a conversation. But those systems have notable limitations. Either they do not incorporate explicit humor algorithms (Zhang et al. 2020), or they only utilize one joke production mechanism (Dybala et al. 2008; Ritschel et al. 2019; Zhu 2019). In contrast, Witscript 3 generates conversational jokes using three joke production mechanisms, which are based on two explicit humor algorithms.

One of those humor algorithms, derived from the Surprise Theory of Laughter (Toplyn 2021), specifies that a monologue-type joke has these three parts: (a) the topic, (b) the angle, and (c) the punch line.

Witscript 3, like Witscript (Toplyn 2021) and Witscript 2 (Toplyn 2022), also incorporates the Basic Joke-Writing Algorithm (Toplyn 2021), which consists of five steps for writing a three-part joke:

1. **Select a topic.** A good joke topic is one sentence that is likely to capture the attention of the audience.
2. **Select two topic handles.** The topic handles are the two words or phrases in the topic that are the most attention-getting.
3. **Generate associations of the two topic handles.** An association is something that the audience is likely to think of when they think about a particular subject.
4. **Create a punch line.** The punch line is the word or phrase that results in a laugh. It links an association of one topic handle to an association of the other topic handle in a surprising way.
5. **Generate an angle between the topic and punch line.** The angle is a word sequence that connects the topic to the punch line in a natural-sounding way.

Now I'll describe how the Witscript 3 system employs those two humor algorithms to improvise jokes in a conversation.

## Description of the Witscript 3 System

Witscript 3 is a neural-symbolic hybrid AI system. It's symbolic because it incorporates the two humor algorithms described above. And it's neural because it executes those algorithms by calling on a transformer-based, large language model (LLM), OpenAI's GPT-3 (Brown et al. 2020). The most capable GPT-3 model currently available, text-davinci-002, is used without fine-tuning. The model has 175 billion parameters and was trained on a filtered version of Common Crawl, English-language Wikipedia, and other high-quality datasets.

Like Witscript and Witscript 2, Witscript 3 uses the two humor algorithms described above to divide the task of improvising a joke into steps. The Witscript 2 and Witscript 3 systems carry out those steps by making a separate call to the LLM for each step and using the output of each step as an input to the next. In their paper, Wu, Terry, and Cai (2022) refer to this process as "prompt chaining," and observe that the step-by-step nature of the process results in a system that is more debuggable, editable, controllable, and explainable than an LLM-based system that accomplishes an entire task all at once.

More specifically, Witscript 3 employs GPT-3 to carry out the steps of the Basic Joke-Writing Algorithm in this way:

1. **Get a topic.** Witscript 3 receives a sentence from a user and treats it as the topic of a three-part joke that consists of a topic, an angle, and a punch line. For example, a user might say to Witscript 3, "Authorities caught two pigs that were wandering around loose in San Antonio, Texas."

2. **Select two topic handles.** The GPT-3 API is called with a prompt to select the two most conspicuous nouns, noun phrases, or named entities in the topic. That's because the humor of human-written jokes tends to be based on nouns and noun phrases (West and Horvitz 2019). From that example topic, GPT-3 selects the topic handles "pigs" and "San Antonio."

3. **Generate associations of the two topic handles.** The GPT-3 API is called with a prompt to generate a list of associations for each topic handle. In our example, for "pigs" GPT-3 generates a list including bacon, pork chops, ham, and sausage. For "San Antonio" it generates a list including The Alamo, River Walk, Texas Longhorns, and Whataburger.

4. **Create three punch line candidates.** Witscript 3 links associations of the topic handles in three different ways to create three punch line candidates: a wordplay candidate, a common-sense knowledge candidate, and a third candidate. To create its **wordplay candidate**, Witscript 3, like Witscript, uses well-known tools of natural language processing to combine one association from each list into a punch line that exhibits wordplay (Toplyn 2020). Witscript 3 does not call on GPT-3 to create its wordplay candidate because GPT-3 seems to be weak at phonetic tasks like generating puns and rhymes, possibly as a result of its use of byte-pair encoding (Branwen 2020). To create its **common-sense knowledge candidate**, Witscript 3, like Witscript 2, uses GPT-3 to combine one association from each list using common-sense knowledge. In our example, when the GPT-3 API is called, GPT-3 combines the associations "sausage" and "The Alamo" into the punch line "Alamo Sausage." Finally, to create its **third candidate**, Witscript 3 uses GPT-3 to power a third, proprietary, joke production mechanism involving the topic handles.

5. **Generate an angle between the topic and each candidate punch line.** The GPT-3 API is called with prompts to generate three joke candidates, each one based on the topic and ending with one of the punch line candidates.

6. **Output the joke candidate that is most likely the funniest.** The GPT-3 API is called with a prompt to determine which of the three joke candidates seems to be the funniest. Then Witscript 3 outputs that joke to the user as its response. In our example, after the user says, "Authorities caught two pigs that were wandering around loose in San Antonio, Texas," Witscript 3 outputs the response, "They were taken to the Alamo Sausage Company."

## System Evaluation

For inputs to evaluate Witscript 3, I used 13 sentences taken from Amazon's Topical-Chat dataset (Gopalakrishnan et al. 2019). The dataset consists of comments exchanged by pairs of workers on Amazon Mechanical Turk (AMT) who were asked to have coherent and engaging conversations based on topical reading material that they had been provided with. The dataset is available from `www.kaggle.com/datasets/arnavsharmaas/chatbot-dataset-topical-chat`.

I took the following steps to select and standardize sentences from the Topical-Chat dataset for use in evaluating Witscript 3:

1. Take a comment from the dataset at random.

2. Select the last (or only) complete sentence in that comment if it meets all of the following criteria: (a) its length is 20 words or less; (b) it has no pronouns whose antecedents are unclear; (c) it has at least two nouns, noun phrases, or named entities; and (d) it isn't basically the same as a sentence that has already been selected. Those are criteria that, in my judgment, a human would use when deciding whether a particular sentence in a conversation could be responded to with a joke relatively easily.

3. Repeat the first two steps until 13 sentences have been selected.

4. Standardize those 13 sentences by correcting any errors in capitalization, spelling, and punctuation.

Then I used each of those 13 sentences as an input to obtain responses from three different sources:

1. **Human**—This is the AMT worker whose response to that sentence was recorded in the Topical-Chat dataset. I took the entire response up until the AMT worker changed the subject. Then I corrected any errors in capitalization, spelling, and punctuation so that the Human responses would be comparable to the responses output by Witscript 3 and GPT-LOL, which contained no such errors.

2. **GPT-LOL**—This is a simple joke generator I created to serve as a baseline. It is the text-davinci-002 version of GPT-3 given the prompt "You want to be funny. Respond to this: [The sentence]." Temperature is set to 0.7 and Top P to 1.0.

3. **Witscript 3**—This is the system described above.

To evaluate the responses produced by the Human, GPT-LOL, and Witscript 3, I hired workers via AMT. I only specified that the AMT workers had to be located in the United States and have a Human Intelligence Task (HIT) Approval Rate greater than or equal to 95%.

The 39 input and response pairs, 13 pairs for each source, were put in a random order and presented in that order to all of the AMT evaluators, who were not told how the pairs had been obtained. Given each input and response pair, the evaluators were asked to rate the response as either 1 (not a joke), 2 (almost a joke), 3 (a joke), or 4 (a very good joke). This evaluation method (Toplyn 2021) is consistent with that of Amin and Burghardt (2020), who use as one criterion for evaluating a humor generation system its humorousness, which they define as the ability of the system to output text that is identifiable as being a joke or as having humorous intent.

Each of the 39 input and response pairs was evaluated by 15 AMT workers, yielding a total of 585 ratings. Table 1 shows the Human's response to each input sentence, along with the first response that was output by the two AI systems for that input. The mean rating given by the evaluators for each response is also shown.

| Source | Input and responses | Mean rating |
| --- | --- | --- |
| | **1. Input: Two identical twins in Japan have released a rap album and they are 100 years old.** | |
| Human | That is awesome. | 2.00 |
| GPT-LOL | This is amazing! I can't believe they are still alive, let alone rapping! | 1.60 |
| Witscript 3 | I'm not sure if they're 'twinning' or 'losing.' | 2.33 |
| | **2. Input: I would love to be the UN ambassador to aliens though!** | |
| Human | Me too lol. That would be a pretty cushy job, I think. | 2.47 |
| GPT-LOL | I think you would make an excellent UN ambassador to aliens! | 1.87 |
| Witscript 3 | I would love to be the US Permanent Representative to the United Aliens! | 2.20 |
| | **3. Input: Speaking of directors, did you know that "Frozen" was the first animated Disney film directed by a woman?** | |
| Human | Yes, "Frozen," for which she earned an Academy Award for Best Animated Feature. Lee is the first female director. | 1.53 |
| GPT-LOL | Yes, I actually did know that! I think it's amazing that Disney is finally starting to represent women in leadership positions! | 1.47 |
| Witscript 3 | And the last. | 1.60 |
| | **4. Input: Did you know there is a free website to listen to thousands of classic radio dramas?** | |
| Human | No, I did not. That sounds pretty cool. | 1.40 |
| GPT-LOL | Yes, I did know that. It's called the Internet Archive. | 1.47 |
| Witscript 3 | Yes, it's called a graveyard. | 1.93 |
| | **5. Input: Germany has given animals legal rights in their constitution.** | |
| Human | I heard about that! I think they started that in 2002. | 1.20 |
| GPT-LOL | This is a hoot! I didn't know that Germany had given animals legal rights in their constitution. I wonder what kinds of rights they have. | 1.60 |
| Witscript 3 | If animals have legal rights, does that mean I can sue my neighbor's dog for barking? | 2.80 |
| | **6. Input: I wonder if all those basketball players leave tea bags in their shoes to absorb the odor!** | |
| Human | It would be hard running the floor with those in there... Just kidding! | 2.60 |
| GPT-LOL | You're hilarious! | 2.13 |
| Witscript 3 | Michael Jordan is the Earl Grey of slam dunks. | 2.67 |
| | **7. Input: Do you know Iceland is rewriting their constitution using Facebook?** | |
| Human | Great way to get more involved via social networking! | 2.13 |
| GPT-LOL | Yes, I heard that Facebook is now the go-to source for constitutional law. | 1.87 |
| Witscript 3 | I'm not surprised. Facebook is where all the cool kids are. | 2.47 |
| | **8. Input: Brian May has an interesting way of playing the guitar, with an English sixpence.** | |
| Human | That's interesting. I heard Brian May has a PhD in astrophysics. | 1.53 |
| GPT-LOL | Brian May has an interesting way of playing the guitar, with an English sixpence. I'm not sure if that's a good thing or a bad thing, but it's certainly unique! | 1.53 |
| Witscript 3 | I always suspected Brian May was a bit of a tightwad! | 2.33 |

Table 1: Responses generated for each input, with their mean ratings on a scale from 1 (not a joke) to 4 (a very good joke).

| | 9. Input: Did you know that panda researchers wear panda costumes to work? | |
|---|---|---|
| Human | That's weird. | 1.73 |
| GPT-LOL | That's a bit of a bamboozle! | 2.93 |
| Witscript 3 | Do they also get a discount at the Panda Express? | 2.53 |
| | 10. Input: Did you know the White House has twin buildings in Ireland and France? | |
| Human | That is interesting. | 1.80 |
| GPT-LOL | Yes, I did know that! The White House is a very popular tourist destination, so it's no surprise that they would have twin buildings in other countries. | 1.73 |
| Witscript 3 | So that's where they've been hiding the other presidents! | 2.93 |
| | 11. Input: If you live in South Africa, you can even attach a flamethrower on your car so it doesn't get carjacked! | |
| Human | That's awesome! I would totally make burgers or something with that flamethrower LOL. But I also probably would not go to South Africa. | 2.07 |
| GPT-LOL | I don't know about you, but I feel safer already! | 2.47 |
| Witscript 3 | I always attach a flamethrower to my car. Just in case I need to light my cigarettes. | 2.13 |
| | 12. Input: An aluminum piano was once built for an airship and weighed only 365 pounds! | |
| Human | "Only," haha. That's still massively heavy. That piano was built for the famous Hindenburg. | 2.00 |
| GPT-LOL | If that's the case, I'd love to see a grand piano made out of aluminum! | 2.07 |
| Witscript 3 | Why did the aluminum piano cross the road? To get to the other pie piano! | 2.13 |
| | 13. Input: There is a radio station that turns solar activity to sound. | |
| Human | Wow, cool. | 1.40 |
| GPT-LOL | Why didn't they just name it "The Sun FM"? | 2.80 |
| Witscript 3 | If you listen to it for too long, you'll get sunburn. | 2.67 |

Table 1 (continued).

Table 2 compares the sources based on their ratings. The second column shows that the Human responses are rated lowest. This is not surprising because the humans recorded in the Topical-Chat dataset were not making any special attempt to be funny. Witscript 3's responses were rated, on average, substantially higher than the responses of GPT-LOL. Those results indicate that a hybrid of an LLM and symbolic humor algorithms results in a higher-quality joke generator than an LLM that has merely been prompted to be funny.

The last column of Table 2 shows the percentage of responses that the evaluators rated as "a joke" or "a very good joke." Witscript 3's responses were judged to be jokes 44% of the time. This result may be evidence of the feasibility of an AI system that can improvise conversational jokes as effectively as a non-expert human can.

| Source | Mean rating | % jokes (rated 3 or 4) |
|---|---|---|
| Human | 1.84 | 23.6% |
| GPT-LOL | 1.96 | 33.8% |
| Witscript 3 | 2.36 | 44.1% |

Table 2: Comparison of the sources based on their ratings.

## Contributions and Future Work

This paper makes the following contributions:

1. It presents an automatic, easily scalable system for improvising a variety of contextually integrated jokes that are based on a variety of joke production mechanisms.
2. It provides evidence that computational humor is best accomplished by taking a hybrid neural-symbolic approach.
3. It demonstrates how an AI system can generate a joke in a series of editable steps, which allows a human to collaborate with the system in creating the joke.
4. It introduces a simple joke generator—GPT-LOL—for use as a baseline in evaluating systems that generate conversational jokes.

To get Witscript 3 to generate jokes more consistently, the following will be explored: using different prompts and configuration settings for GPT-3; using LLMs other than GPT-3 to execute the humor algorithms; and incorporating additional joke production mechanisms to widen the variety of the joke outputs even more.

Work will also be devoted to developing a hybrid of an LLM and the humor algorithms described herein that can recognize jokes in addition to generating them.

## Conclusion

The Witscript 3 system could be integrated into a chatbot as a humor module; the proprietary software is available for license. Such a humor-enabled chatbot might animate an artificial, but likeable, companion for lonely humans.

# References

Amin, M., and Burghardt, M. 2020. A Survey on Approaches to Computational Humor Generation. In Proceedings of the 4th Joint SIGHUM Workshop on Computational Linguistics for Cultural Heritage, Social Sciences, Humanities and Literature, 29–41. Online: International Committee on Computational Linguistics.

Branwen, G. 2020. GPT-3 Creative Fiction. www.gwern.net/GPT-3#bpes. Accessed: 2022-10-24.

Brown, T. B.; Mann, B.; Ryder, N.; Subbiah, M.; Kaplan, J.; Dhariwal, P.; Neelakantan, A.; Shyam, P.; Sastry, G.; Askell, A.; Agarwal, S.; Herbert-Voss, A.; Krueger, G.; Henighan, T. J.; Child, R.; Ramesh, A.; Ziegler, D. M.; Wu, J.; Winter, C.; Hesse, C.; Chen, M.; Sigler, E.; Litwin, M.; Gray, S.; Chess, B.; Clark, J.; Berner, C.; McCandlish, S.; Radford, A.; Sutskever, I.; and Amodei, D. 2020. Language Models are Few-Shot Learners. ArXiv, abs/2005.14165.

Dybala, P.; Ptaszynski, M.; Higuchi, S.; Rzepka, R.; and Araki, K. 2008. Humor Prevails! - Implementing a Joke Generator into a Conversational System. In Wobcke, W.; and Zhang, M., eds., AI 2008: Advances in Artificial Intelligence (AI 2008). Lecture Notes in Computer Science, vol. 5360, 214–225. Berlin, Heidelberg: Springer. doi.org/10.1007/978-3-540-89378-3_21.

Gopalakrishnan, K.; Hedayatnia, B.; Chen, Q.; Gottardi, A.; Kwatra, S.; Venkatesh, A.; Gabriel, R.; and Hakkani-Tür, D. 2019. Topical-Chat: Towards Knowledge-Grounded Open-Domain Conversations. In Proceedings of Interspeech 2019, 1891-1895, doi.org/10.21437/Interspeech. 2019-3079.

Ritchie, G. 2005. Computational Mechanisms for Pun Generation. In Wilcock, G.; Jokinen, K.; Mellish, C.; and Reiter, E., eds., Proceedings of the 10th European Natural Language Generation Workshop, 125-132. Morristown, NJ, USA: ACL Anthology.

Ritschel, H.; Aslan, I.; Sedlbauer, D.; and André, E. 2019. Irony Man: Augmenting a Social Robot with the Ability to Use Irony in Multimodal Communication with Humans. In Proceedings of the 18th International Conference on Autonomous Agents and Multi-Agent Systems (AAMAS '19), 86–94. Richland, SC, USA: International Foundation for Autonomous Agents and Multiagent Systems.

Toplyn, J. 2020. Systems and Methods for Generating Comedy. U.S. Patent No. 10,878,817. Washington, DC, USA: U.S. Patent and Trademark Office.

Toplyn, J. 2021. Witscript: A System for Generating Improvised Jokes in a Conversation. In Proceedings of the 12th International Conference on Computational Creativity, 22–31. Online: Association for Computational Creativity.

Toplyn, J. 2022. Witscript 2: A System for Generating Improvised Jokes Without Wordplay. In Proceedings of the 13th International Conference on Computational Creativity, 54-58. Online: Association for Computational Creativity.

West, R., and Horvitz, E. 2019. Reverse-Engineering Satire, or "Paper on Computational Humor Accepted Despite Making Serious Advances." In Proceedings of the Thirty-Third AAAI Conference on Artificial Intelligence and Thirty-First Innovative Applications of Artificial Intelligence Conference and Ninth AAAI Symposium on Educational Advances in Artificial Intelligence (AAAI'19/IAAI'19/EAAI'19), Article 892, 7265–7272. Palo Alto, CA, USA: AAAI Press. doi.org/10.1609/aaai.v33i01.33017265.

Winters, T. 2021. Computers Learning Humor Is No Joke. *Harvard Data Science Review*, *3*(2). doi.org/10.1162/99608f92.f13a2337.

Wu, T.; Terry, M.; and Cai, C. J. 2022. AI Chains: Transparent and Controllable Human-AI Interaction by Chaining Large Language Model Prompts. In Proceedings of the 2022 CHI Conference on Human Factors in Computing Systems (CHI '22), Article 385, 1–22. New York, NY, USA: Association for Computing Machinery. doi.org/10.1145/3491102.3517582.

Zhang, H.; Liu, D.; Lv, J.; and Luo, C. 2020. Let's be Humorous: Knowledge Enhanced Humor Generation. arXiv preprint. arXiv:2004.13317.

Zhu, D. 2019. Humor robot and humor generation method based on big data search through IOT. Cluster Computing, 22: 9169–9175. doi.org/10.1007/s10586-018-2097-z.